%% file: main.tex
\documentclass[a4paper]{article}

\usepackage{INTERSPEECH2020}
\usepackage[table]{xcolor}
\usepackage[colorinlistoftodos]{todonotes}
\usepackage{multirow}
\usepackage{amsmath}
\usepackage{float}
\usepackage{graphicx}
\usepackage[ruled,vlined]{algorithm2e}
\usepackage{subcaption}
\usepackage{url}
\usepackage{todonotes}

\newcommand{\etal}{\emph{et al}}

\newcommand{\argmin}[1]{\underset{#1}{\operatorname{arg}\,\operatorname{min}}\;}

\SetCommentSty{mycommfont}

\SetKwInput{KwInput}{Input}                
\SetKwInput{KwOutput}{Output}              

\title{Representation based meta-learning for few-shot spoken intent recognition}

\name{Ashish Mittal, Samarth Bharadwaj, Shreya Khare, Saneem Chemmengath,\\ Karthik Sankaranarayanan, Brian Kingsbury}
\address{
  IBM Research AI
  }
\email{\{arakeshk,samarthb,skhare34,saneem.cg,kartsank\}@in.ibm.com, bedk@us.ibm.com}

\begin{document}

\maketitle

\begin{abstract}
Spoken intent detection has become a popular approach to interface with various smart devices with ease. However, such systems are limited to the preset list of \emph{intents-terms} or \emph{commands}, which restricts the quick customization of personal devices to new intents. This paper presents a few-shot spoken intent classification approach with task-agnostic representations via meta-learning paradigm. Specifically, we leverage the popular representation based meta-learning learning to build a task-agnostic representation of utterances, that then use a linear classifier for prediction. We evaluate three such approaches on our novel experimental protocol developed on two popular spoken intent classification datasets: Google Commands and the Fluent Speech Commands dataset. For a 5-shot (1-shot) classification of novel classes, the proposed framework provides an average classification accuracy of 88.6\% (76.3\%) on the Google Commands dataset, and 78.5\% (64.2\%) on the Fluent Speech Commands dataset. The performance is comparable to traditionally supervised classification models with abundant training samples.
\end{abstract}

\section{Introduction}

\input{intro}

\begin{figure}[!t]
    \centering
    \includegraphics[width=0.45\textwidth]{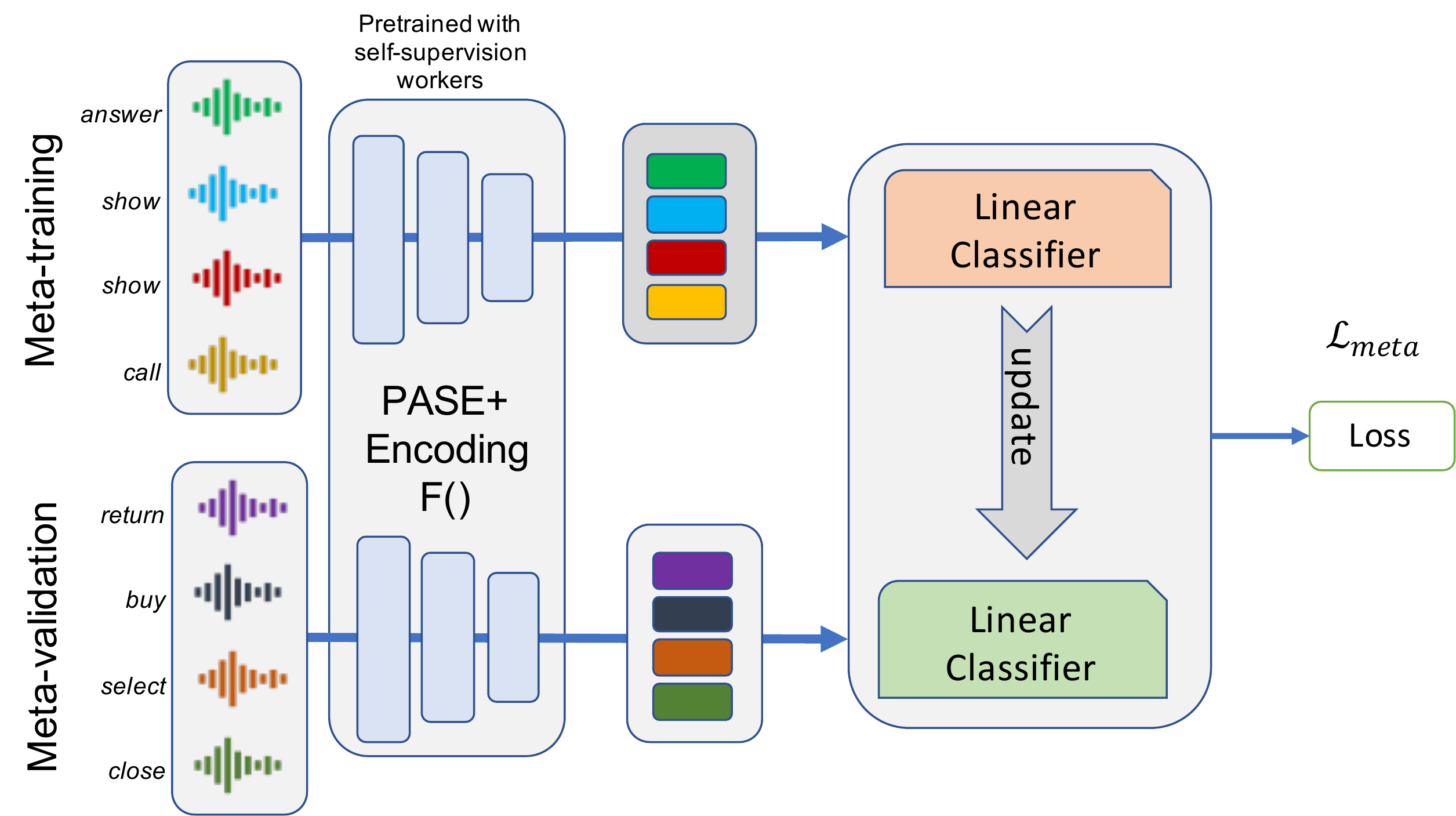}
    \caption{The representation based few shot encoding learns a task agnostic feature representation. Simultaneously, the accompanying linear classifier (nearest-neighbour, SVM) learns tasks-specific class discrimination boundaries. Therefore, the proposed approach can adapt to a new task with few samples (few-shot) by updating just the linear classifier.}
    \label{fig:proposed}
\end{figure}

\section{Proposed approach}

As shown in Figure \ref{fig:proposed}, the proposed representation based few-shot learning paradigm is trained to be able to produce a task-agnostic representation of an utterance using the popular meta-learning paradigm. In this work, we define a \emph{task} or \emph{episode} as a $n$-\emph{way}, $m$-\emph{shot} classification problem, i.e., $n$ classes are predicted with only (small) $m$ audio samples per class available for training. In this section, we describe the meta-learning pipeline for few-shot learning. 

\subsection{Meta-learning}
Task independent learning, or learning to learn is a popular paradigm of machine learning termed meta-learning  \cite{vinyals2016matching, finn2017modelagnostic, lee2019meta, bertinetto2018metalearning, snell2017prototypical, gidaris2018dynamic}. 
We consider three popular approaches from the \emph{robust-representation} view of the meta-learning paradigm, termed ProtoNet \cite{ snell2017prototypical}, Ridge \cite{bertinetto2018metalearning} and MetaOptNet \cite{lee2019meta}. The representation view achieves the goal of few-shot learning by decoupling the learning process into: task-independent encoding function $\mathcal{F}_\theta$ parameterized by $\theta$,
and task-specific classifier $\mathcal{A}_\phi$ parameterized by $\phi$. In general, $\mathcal{F}_\theta$ is a complex neural network model and $\mathcal{A}_\phi$ a simple linear classifier that project the embedding from $\mathcal{F}_\theta$ to space of class labels.
The optimizer minimizes the expectation of \emph{meta}-loss $\mathcal{L}^{meta}$ (negative log-likelihood) over $T$ tasks to obtain task-independent optimal parameters $\theta$ (as given by Equation \ref{eq:metaloss}).
\begin{equation}
\label{eq:metaloss}
\begin{split}
\min_{\theta} \frac{1}{T} \sum^{T}_{t=1}\mathcal{L}^{meta}(\mathcal{D}^{t}_{q};  \mathcal{A}_{\phi},\mathcal{F}_\theta) + \mathcal{R}(\theta) \\
\text{where } \phi=\argmin{\phi} \ell(\mathcal{D}^{t}_{s}; \mathcal{A}_\phi,\mathcal{F}_\theta)
\end{split}
\end{equation}

\begin{algorithm}
\scriptsize
\DontPrintSemicolon
\caption{\label{algo:batch} Meta-training routine for single batch of $n$-\emph{way}, $m$-\emph{shot} classification}

  \KwInput{$\mathcal{F}(),\mathcal{A},crossEntropy(), generateEpisode()$ \cite{vinyals2016matching}}
  \KwOutput{$\theta^{\star}$}
  \KwData{$\mathcal{D}_{tr}$} 
  \tcc{$n$ classes are chosen from $y$, for which $m$ samples are chosen as \emph{support}, while $m$ from remaining samples are chosen as \emph{query}}
  
  i=0 loss=list()
  
  \For{i $\leq$ episodes\_per\_batch}
  {
  support, supportLabels, query, queryLabels = generateEpisode($\mathcal{D}_{tr}$, m, n)
  \tcc{Sample n-\emph{way}, m-\emph{shot} episode from $\mathcal{D}_{tr}$}
  
  $E_{support}=\mathcal{F}(x_{support})$ 
  
  $E_{query}=\mathcal{F}(x_{query})$ 
  
  $\mathcal{A}$.\emph{train}($E_{support}$, supportLabels)
  
  $logits$ = $\mathcal{A}$.predict($E_{query}$)
  
  $\mathcal{L}^{meta}$ = $crossEntropy$(logits, queryLabels)
  
  loss.\emph{append}($\mathcal{L}^{meta}$)
}
 
 $\theta^{\star}$=optimizer.backward(mean(loss))
 \tcc{mean classification loss across all episodes of a batch is backprop}
\end{algorithm}

\noindent where $[(\mathcal{D}^t_s, \mathcal{D}^t_q)]^{T}_{t=1}$, 
is the few-shot episode generated from training set which are n-\emph{way}, m-\emph{shot} classification tasks, following the strategy proposed in \cite{vinyals2016matching}. Every episode has a support set $\mathcal{D}^t_s$ which is used to train the task classifier $\mathcal{A}_\phi$ and and query set $\mathcal{D}^t_q$ to train $\mathcal{F}_\theta$. 
Further, $\ell$ is the task-classification loss which depends on the classifier $\mathcal{A}_\phi$ used. Task classifier could be a nearest neighbour (ProtoNet \cite{snell2017prototypical}), linear regression (Ridge \cite{bertinetto2018metalearning}) or linear SVM (MetaOptNet \cite{lee2019meta}).
In order to train the entire setup in an end-to-end fashion, in case of SVM, implicit functions are utilized to capture \emph{approximate} gradients to train the embedding function \cite{lee2019meta}. During the validation and test routines, similar episodes are sampled from a \emph{novel} set of classes that are previously unseen by the classifier. However, only the task classifier ($\mathcal{A}$) is updated from the validation or test episode samples. The meta-learning loss ($\mathcal{L}^{meta}$) is weight regularized ($\mathcal{R}(\theta)$) to ensure generalization and preventing over-fitting of the network. Hence, the batch optimization is performed over multiple episodes to compute the average $\mathcal{L}^{meta}$. The details of the encoder embedding function is described next. 

\subsection{Self-supervised encoder}
For the input utterances encoder ($\mathcal{F}(.)$), we use a convolutional neural network as the embedding function. Specifically, we utilize the PASE+ architecture \cite{pascual2019learning, ravanelli2020multitask} pre-trained on LibriSpeech dataset\cite{panayotov2015librispeech} with self-supervised tasks obtained from the original signal. The convolution based architecture first encodes the input with SincNet \cite{ravanelli2018speaker}, that learns custom band-pass filters for the input signal. Next, seven residual network blocks are applied with skip connections \cite{he2015deep} that are pooled together with the QRNN temporal pooling \cite{bradbury2016quasi} method. Additionally, in this work, we propose to continue to regularize the embedding learner obtained from PASE+ with some of the self-supervision tasks that have been shown to improve representation power of the embedding. We utilize classification \emph{workers} that reconstruct static-features engineered from the input speech signal from the same base encoding used for meta-learning. To the best of our knowledge, our work is first to explore the benefits of self-supervision in conjuncture with meta-learning task. We also hypothesize a weighted combination of the reconstruction loss from the self-supervision (with a controlling parameter $\alpha$) together with the cross-entropy loss from the meta-learning few-shot classification, as $\mathcal{L}^{total}=\alpha \mathcal{L}^{meta} + (1-\alpha) \sum^{W}_{w=0} \mathcal{L}^{self}_{w}$, where $\mathcal{L}^{self}$ is the combined loss of all workers of PASE+.
The overall routine of the proposed approach for a single batch is shown in Algorithm \ref{algo:batch}. As mentioned, PASE+ which is pre-trained with only self-supervision is used as the embedding function. Next, we show the empirical evaluation of the proposed architecture for few-shot learning. We showcase $5$-shot an $1$-shot classification of intent on two popular datasets. 

\input{exp}

\section{Conclusions}
In this work, we show the classification and generalization of our few-shot learning with the representation based meta-learning paradigm. On two benchmarks for spoken intent classification, we show that the proposed approach can fine-tune with limited examples (we tested one and five) of novel classes and perform classification task with sufficient accuracy. The proposed approach also benefits from the self-supervised pretrained embedding function trained on an additional dataset. Intent classification and command detection have been prevalent in several consumer devices. However, the list of commands have as yet been static or pre-programmed. Our results indicates the possibility of a light weight classification model (such as a nearest neighbor or linear SVM) that can be easily fine-tuned for new commands with limited training examples. 



\bibliographystyle{IEEEtran}
\bibliography{mybib}
\end{document}

%% file: intro.tex
Recent advances in AI powered consumer devices has brought forth huge advances in direct speech to intent or command detection. Unlike traditional speech problems, such as speech to text, which are designed to weigh all words equally, the correct prediction of intent requires identifying the essential tokens or phrases correctly. Spoken command detection is a challenging problem since the utterance may be unclear, intent may be implicit, and the response time must be low. In such a scenario, advances made in natural language understanding can be rarely leveraged and the intent or command must be identified directly from speech. 

Kao \etal \cite{Kao_2019} showcase the low memory footprint and computational time of utilizing convolution network for the task of keyword spotting, reporting over $97$\% accuracy on the Speech Commands dataset \cite{speechcommands}. Recently, Poncelet \etal \cite{Poncelet_2020} leverage the advances in capsules network architecture for multi-task prediction of speaker and label of command, showing the robustness of the model. Further, some approaches leverage additional text data to improve performance of speech to intent classification in low resource settings \cite{upairedtext}. 

\begin{figure}[!t]
    \centering
    \includegraphics[width=0.45\textwidth]{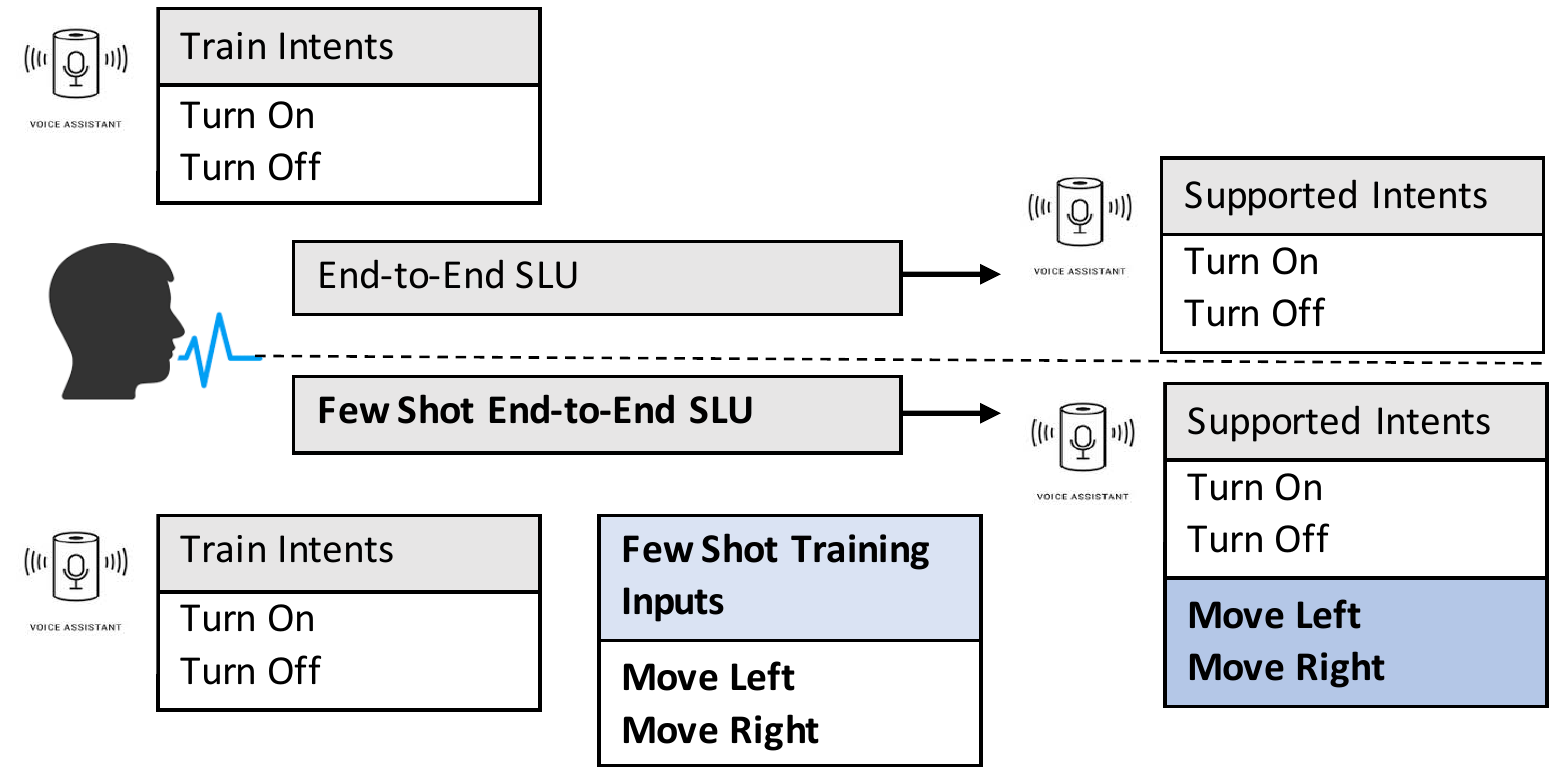}
    \caption{An illustration of the proposed few-shot approach to End-to-End SLU. A user adds new commands `move' to the voice assistant.}
    \label{fig:intro}
\end{figure}

Similarly, several other end-to-end learning methods have been proposed for spoken language understanding (SLU) task ~\cite{Qian2017ExploringAE,8639043,endtoend1,endtoend3,8461718,endSahar,upairedtext,sig2sem,Curriculumlearning,pretrainingASR,Bhosale2019EndtoEndSL} without requiring intermediate text and show promising results on multiple tasks. However, their success heavily depends on a large amount of labeled training data. Such data is usually scarce and thus limiting the performance. Furthermore, such approaches limit personal devices to a specific list of commands or intents, restricting customization. This is illustrated in  Figure \ref{fig:intro}. This paper brings the meta-learning paradigm to few-shot speech to intent classification task. 

We show on two popular spoken intent detection datasets: the Google Speech Commands \cite{speechcommands} and Fluent Speech Commands \cite{pretrainingASR} datasets, that $1$-shot and $5$-shot classification to predict spoken commands and intents can be performed with surprisingly high accuracy. Specifically, we show that a representation based meta-learning paradigm that first learns task-agnostic representation of input utterance followed by task-specific base learners are ideal for spoken-command feature that is popular in several consumer devices. Representation based meta-learning (R-ML) simultaneously leverage the representational power of neural networks, with the low-resource requirements of linear classifiers. Further, the models are trained in an end-to-end fashion thereby passing gradients computed from classification error through the linear classifier and the representational neural network simultaneously. The end-to-end nature of these architectures makes them powerful \emph{generalizers}, in which the small foot-print linear classifier can be easily retrained with the few-shot examples available of novel classes. 

To ensure that such a classifier ensemble does not overfit to few-shot available per class, we utilize a speech representation known as PASE+ \cite{pascual2019learning, ravanelli2020multitask}, that is trained with only self-supervised tasks, i.e., classification tasks generated from the data itself, such as reconstruction of static-engineered features. We hypothesize that the additional strong regularization of self-supervision ensures the representation remains task-agnostic. To the best of our knowledge, this is the first study to evaluate three popular representation based meta-learning approaches for speech classification tasks as well as self-supervised regularization of meta-learning. 

Similar to our proposed approach, \cite{chen2018investigation} present a meta-learning formulation of intent classification problem on the Google commands dataset. Different from R-ML, the performance of a popular optimization based approach to meta-learning, known as model-agnostic-meta-learning (MAML) \cite{finn2017modelagnostic} is evaluated on small subsets of the dataset, though MAML has shown to be difficult to train in practice \cite{rajeswaran2019metalearning}. On the other hand, Koluguri \etal \cite{meta-adult} show that meta-learning with ProtoNet \cite{snell2017prototypical} is effective for child-adult speaker diarization by modeling each speaker session as novel task. Neither works perform a thorough evaluation of multiple meta-learning algorithms on standard protocols of command or intent classification problems.

%% file: exp.tex
\section{Experiments}

\subsection{Dataset}

Experiments are performed on two popular datasets: Google Speech Commands dataset \cite{speechcommands} and Fluent Speech Commands\cite{pretrainingASR}. 
This work presents an speech to intent classification task with few-shots per label, the number of samples per class and classes seen in train and test is controlled via a novel protocol first proposed in ~\cite{vinyals2016matching}. First, we split the datasets into train, validation and test such that all labels are novel (completely different class labels in each split). The details of distribution in the revised protocol resulting from the new splits are shown in Table~\ref{tab:dataset}. Additionally, we prune the splits by utilizing the available speaker labels to create two versions of the meta-learning protocol, with and without speaker overlap, referred to as SPO and No-SPO, respectively.\footnote{The dataset splits used in our paper are available at \url{https://github.com/AshishMittal/RMLIntent}}.
A similar protocol is proposed by \cite{chen2018investigation}, however, the authors ignore speaker overlap that may inadvertently boost the reported performance.

\begin{table}[!hb]
\caption{\label{tab:dataset} Dataset statistics on Google Commands and Fluent Speech Commands of the proposed meta-learning protocol}
\resizebox{\columnwidth}{!}{%
\begin{tabular}{|l|l|l|l|l|l|l|l|}
\hline
\multicolumn{2}{|l|}{\multirow{2}{*}{}} &
  \multicolumn{3}{c|}{Google Commands} &
  \multicolumn{3}{c|}{Fluent Speech Commands} \\ \cline{3-8} 
\multicolumn{2}{|l|}{}                    & Train  & Validation & Test   & Train  & Validation & Test  \\ \hline
                       & \#Classes        & 18     & 5          & 5      & 15     & 8          & 8     \\ \hline
\multirow{3}{*}{SPO}   & \#Audio Files    & 42,954 & 10,560     & 11,207 & 13,030 & 7,632      & 8,572 \\ \cline{2-8} 
                       & \#Speakers       & 1881   & 1881       & 1881   & 97     & 97         & 97    \\ \cline{2-8} 
                       & Duration (hours) & 8.28   & 4.85       & 5.27   & 10.44  & 3.54       & 3.72  \\ \hline
\multicolumn{1}{|c|}{\multirow{3}{*}{\begin{tabular}[c]{@{}c@{}}No- \\  SPO\end{tabular}}} &
  \#Audio Files &
  30,181 &
  1,389 &
  1,436 &
  10,115 &
  811 &
  982 \\ \cline{2-8} 
\multicolumn{1}{|c|}{} & \#Speakers       & 1503   & 189        & 189    & 77     & 10         & 10    \\ \cline{2-8} 
\multicolumn{1}{|c|}{} & Duration (hours) & 6.39   & 0.5        & 0.65   & 8.24   & 0.38       & 0.39  \\ \hline
\end{tabular}%
}
\end{table}

\begin{table}[!t]
\caption{Test Intent Accuracy comparison of the ProtoNet\cite{snell2017prototypical} model with baseline (supervised n-way m-shot) and skyline (supervised n-way all-shot).}
\resizebox{\columnwidth}{!}{%
\begin{tabular}{|l|c|l||c|l|}
\hline
\multicolumn{1}{|c|}{\multirow{2}{*}{Model}}                & \multicolumn{2}{c||}{Google Commands}             & \multicolumn{2}{c|}{Fluent Speech Commands}      \\ \cline{2-5} 
\multicolumn{1}{|c|}{} & 5-shot                         & \multicolumn{1}{c||}{1-shot} & 5-shot                           & \multicolumn{1}{c|}{1-shot} \\ \hline
Baseline               & \multicolumn{1}{l|}{26.22 ± 7} & 22.39 ± 13                  & \multicolumn{1}{l|}{27.42 ± 7} & 21.32 ± 13                  \\ \hline
Skyline                & \multicolumn{2}{c||}{95.61}                                   & \multicolumn{2}{c|}{88.98}                                     \\ \hline
\begin{tabular}[c]{@{}l@{}} \bf ProtoNet\\ \bf \cite{snell2017prototypical}\end{tabular} & \multicolumn{1}{l||}{\bf 78.86 ± 0.40} & \bf 69.30 ± 0.54 & \multicolumn{1}{l|}{\bf 78.00 ± 0.39} & \bf 64.24 ± 0.52 \\ \hline
\end{tabular}
}
\label{tab:result}
\end{table}

\subsection{Design choices}
\noindent {\bf Experiment design}: We show two scenarios of few-shot learning - (i) $5$-way $5$-shot and (ii) $5$-way $1$-shot classification for both validation and test scenarios. Further, we observe empirically that the performance on the test set is better when the number of shots in training episodes are higher, hence we generate $15$-shot episodes in train. All experiments are conducted for a maximum of $30$ epochs, where for each epoch, one batch consists of $1000$ train episodes. The episode generation strategy is performed with replacement, hence an infinite number of episodes may be generated from the training set. We cap the number of episodes per batch based on the capacity of the GPU memory. We sample 2000 episodes from validation and test and report mean and standard deviation of the total n-\emph{way} accuracy \cite{vinyals2016matching}. All the models are optimized using SGD with Nesterov momentum\cite{nesterov1983method}, using an initial learning rate of $0.01$ and momentum of $0.9$ with step decay at every 20 epochs. We use mini-batches of $75$ waveforms of $1$s for Google Commands and $4$s for Fluent Speech Commands dataset. 

\noindent {\bf Embedding}: As mentioned before, we use the PASE+ architecture as the encoder for the input utterances. To ensure the fast performance of the base learners, the original PASE+ encoding of 256 per frame is reduced to a vector size of 50 per frame. Further, a single vector representation is obtained by computing mean across frames. Importantly, our empirical analysis shows that retaining the self-supervised workers does not significantly affect the performance of the downstream task of few-shot classification. We experimented with several loss weighting strategies (by tuning $\alpha$) and found that the self-supervised workers deter performance. We hypothesize that since the meta-learning framework gathers meta-loss $\mathcal{L}^{meta}$ as the average of classification loss across multiple episodes, the model is already sufficiently regularized without the aid of self-supervised workers. However, the PASE+ encoder is pretrained with self-supervision workers and significantly affect the performance, showing that the workers helped to generalize the representation during pretraining and facilitating transfer to the new classes.

\noindent {\bf Benchmark}: To the best of our knowledge, this is the first work to evaluate any approach to few-shot speech to intent classification tasks in literature. Hence, we also set a baseline and a hypothetical skyline on the same experimental protocol:  
\noindent \textbf{Baseline}:  For the baseline, we create a $5$-way $1$-shot and $5$-way $5$-shot intent classification task, where we train the classification model (using the same pretrained PASE+ embedding) for the same $5$ classes as the proposed protocol. 
\noindent \textbf{Skyline:} For the skyline, we create a $5$-way intent classification task, using \emph{all} the training data available for only those classes in the original split of both datasets (on average, $n$ is 1786 in Google commands and 845 in Fluent.ai). Here, the model is tested on the same classes to give us the \emph{maximum} ceiling that can be reached in a traditional classification task. 

\noindent In both cases, the architecture is a traditional multi-layer perceptron network with embedding obtained from the PASE encoder trained with categorical cross-entropy loss. For both the baseline and the skyline we train the encoder, as well as the MLP neural network using SGD.

\begin{table*}[!t]
\centering
\caption{\label{tab:detailed_results} Detailed evaluation on both datasets with three base learners. The no-SPO is the subset of protocol where the speakers do not overlap.}
\resizebox{\textwidth}{!}{%
\begin{tabular}{|c|l|l|l|l|l|l|l|l|l|}
\hline
\multicolumn{2}{|c|}{} &
  \multicolumn{4}{c|}{Google Commands} &
  \multicolumn{4}{c|}{Fluent Speech Commands} \\ \cline{3-10} 
\multicolumn{2}{|c|}{} &
  \multicolumn{2}{c|}{5-Shot} &
  \multicolumn{2}{c|}{1-Shot} &
  \multicolumn{2}{c|}{5-Shot} &
  \multicolumn{2}{c|}{1-Shot} \\ \cline{3-10} 
\multicolumn{2}{|c|}{\multirow{-3}{*}{Model}} &
  \multicolumn{1}{c|}{SPO} &
  \multicolumn{1}{c|}{No-SPO} &
  \multicolumn{1}{c|}{SPO} &
  \multicolumn{1}{c|}{No-SPO} &
  \multicolumn{1}{c|}{SPO} &
  \multicolumn{1}{c|}{No-SPO} &
  \multicolumn{1}{c|}{SPO} &
  \multicolumn{1}{c|}{No-SPO} \\ \hline
 &
  Val &
  {\color[HTML]{000000} 85.45 ± 0.32} &
  {\color[HTML]{000000} 83.48 ± 0.35} &
  {\color[HTML]{000000} 71.07 ± 0.52} &
  {\color[HTML]{000000} 70.61 ± 0.53} &
  {\color[HTML]{000000} 83.55 ± 0.38} &
  {\color[HTML]{000000} 68.57 ± 0.45} &
  {\color[HTML]{000000} 70.42 ± 0.55} &
  {\color[HTML]{000000} 56.14 ± 0.56} \\ \cline{2-10} 
\multirow{-2}{*}{ProtoNet \cite{snell2017prototypical}} &
  \cellcolor[HTML]{C0C0C0}Test &
  \cellcolor[HTML]{C0C0C0}{\color[HTML]{000000} \textbf{89.63 ± 0.27}} &
  \cellcolor[HTML]{C0C0C0}{\color[HTML]{000000} 78.86 ± 0.40} &
  \cellcolor[HTML]{C0C0C0}{\color[HTML]{000000} 74.35 ± 0.50} &
  \cellcolor[HTML]{C0C0C0}{\color[HTML]{000000} 69.30 ± 0.54} &
  \cellcolor[HTML]{C0C0C0}{\color[HTML]{000000} \textbf{78.86 ± 0.40}} &
  \cellcolor[HTML]{C0C0C0}{\color[HTML]{000000} 78.00 ± 0.39} &
  \cellcolor[HTML]{C0C0C0}{\color[HTML]{000000} \textbf{65.61 ± 0.52}} &
  \cellcolor[HTML]{C0C0C0}{\color[HTML]{000000} \textbf{64.24 ± 0.52}} \\ \hline
 &
  Val &
  {\color[HTML]{000000} 82.38 ± 0.34} &
  {\color[HTML]{000000} 81.33 ± 0.35} &
  {\color[HTML]{000000} 68.49 ± 0.51} &
  {\color[HTML]{000000} 68.03 ± 0.51} &
  {\color[HTML]{000000} 81.47 ± 0.39} &
  {\color[HTML]{000000} 67.75 ± 0.42} &
  {\color[HTML]{000000} 70.56 ± 0.51} &
  \cellcolor[HTML]{FFFFFF}{\color[HTML]{000000} 56.69 ± 0.55} \\ \cline{2-10} 
\multirow{-2}{*}{Ridge \cite{bertinetto2018metalearning}} &
  \cellcolor[HTML]{C0C0C0}Test &
  \cellcolor[HTML]{C0C0C0}{\color[HTML]{000000} 89.11 ± 0.29} &
  \cellcolor[HTML]{C0C0C0}{\color[HTML]{000000} 86.17 ± 0.31} &
  \cellcolor[HTML]{C0C0C0}{\color[HTML]{000000} \textbf{75.05 ± 0.48}} &
  \cellcolor[HTML]{C0C0C0}{\color[HTML]{000000} \textbf{76.34 ± 0.49}} &
  \cellcolor[HTML]{C0C0C0}{\color[HTML]{000000} 75.75 ± 0.40} &
  \cellcolor[HTML]{C0C0C0}{\color[HTML]{000000} \textbf{78.51 ± 0.39}} &
  \cellcolor[HTML]{C0C0C0}{\color[HTML]{000000} 61.64 ± 0.53} &
  \cellcolor[HTML]{C0C0C0}{\color[HTML]{000000} 60.84 ± 0.51} \\ \hline
 &
  Val &
  \cellcolor[HTML]{FFFFFF}{\color[HTML]{000000} 76.31 ± 0.39} &
  {\color[HTML]{000000} 83.48 ± 0.35} &
  \cellcolor[HTML]{FFFFFF}{\color[HTML]{000000} 62.90 ± 0.51} &
  {\color[HTML]{000000} 60.50 ± 0.51} &
  \cellcolor[HTML]{FFFFFF}{\color[HTML]{000000} 78.52 ± 0.43} &
  {\color[HTML]{000000} 65.90 ± 0.46} &
  \cellcolor[HTML]{FFFFFF}{\color[HTML]{000000} 67.68 ± 0.53} &
  {\color[HTML]{000000} 52.77 ± 0.53} \\ \cline{2-10} 
\multirow{-2}{*}{MetaOptNet \cite{lee2019meta}} &
  \cellcolor[HTML]{C0C0C0}Test &
  \cellcolor[HTML]{C0C0C0}{\color[HTML]{000000} 84.37 ± 0.32} &
  \cellcolor[HTML]{C0C0C0}{\color[HTML]{000000} \textbf{88.64 ± 0.29}} &
  \cellcolor[HTML]{C0C0C0}{\color[HTML]{000000} 66.61 ± 0.53} &
  \cellcolor[HTML]{C0C0C0}{\color[HTML]{000000} 70.82 ± 0.50} &
  \cellcolor[HTML]{C0C0C0}{\color[HTML]{000000} 70.16 ± 0.43} &
  \cellcolor[HTML]{C0C0C0}{\color[HTML]{000000} 69.84 ± 0.44} &
  \cellcolor[HTML]{C0C0C0}{\color[HTML]{000000} 58.71 ± 0.50} &
  \cellcolor[HTML]{C0C0C0}{\color[HTML]{000000} 59.44 ± 0.51} \\ \hline
\end{tabular}%
}
\end{table*}

\subsection{Evaluation}

In this section we evaluate the performance of the proposed model for both the datasets. As shown in Table \ref{tab:result}, we report accuracy scores as the mean and standard deviation of the accuracy of $2000$ randomly sampled episodes from validation and test splits \cite{vinyals2016matching}. Table \ref{tab:detailed_results} shows a more detailed analysis of various base learners and their performance on SPO protocol and the no-SPO subset. In all cases, the test accuracy is reported at best validation accuracy obtained during the training process.

\noindent As expected, the baseline model performance is close to chance when trained on the proposed few-shot protocol for both the datasets. Note that the baseline is unable to leverage the pretrained PASE+ vectors with such few training samples. On the other hand, the fully supervised skyline (or all-shot experiment) is trained with all the available training data in an end-to-end fashion. While the performance of the model is high, the fine-tuning of the proposed approach only requires fine-tuning of only the base learner, which is a fraction of the computation cost of the skyline training process.

As shown in Table \ref{tab:result}, the proposed approach achieves $76.34$±$0.49$\% and $64.24$±$0.52$\% on the 5-way 1-shot experiment, i.e., on a previously unseen class, the base learner fine-tunes with 5 samples (1 example per class) and provides the mentioned average performance. The low variance across 2000 randomly selected episodes from validation and from test show that the performance is fairly stable across different possible combinations of classes and instances in the test set.  

Next, we show detailed performance analysis of the proposed model with various base learners used in the representational meta-learning framework in Table \ref{tab:detailed_results}. We specifically run each experiment setting with the three popular representation learning protocols - ProtoNet \cite{snell2017prototypical}, Ridge \cite{bertinetto2018metalearning} and MetaOptNet (SVM) \cite{lee2019meta}. The best performing model on the Google Commands dataset for 5-shot classification achieves the accuracy of $89.63$\% at the test time, which is only few percent short of the skyline accuracy of $95.61$\% using only 5 examples in the training for each class. The best accuracy of $88.64$\% on the harder No-SPO Google Commands dataset is achieved using SVM base learner. This result signifies that our model performs equally well when there are new speakers and classes at the test time. Self-supervised learning of the PASE+ encoder and the subsequent representation learning of MetaOptNet, makes sure that the model do not overfit on speaker traits, but instead learn robust representation that is useful for the classification task. The best results for the 1-shot classification on the Google Commands dataset for both SPO and No-SPO variants are $\sim75$\% and $\sim76$\%, which demonstrates the efficacy of the proposed model. 

The best accuracy on the Fluent Speech Commands dataset for $5$-shot classification task is around $\sim79$\% for the SPO variant, which is approximately close to the No-SPO variant. The best accuracy for the 1-shot classification task for the Fluent Speech Commands dataset is around $\sim65\%$ for both the variant of the dataset. There is a drop in the accuracy for the Fluent Speech Commands dataset as compared to Google Commands, because the former has longer utterances($\sim3s$)  as compared to the latter ($\sim1s$). Another reason being that the classes in Fluent Speech Commands overlap significantly (e.g, deactivate lights, deactivate lights kitchen, deactivate lights washroom). Finally,we note that for the SPO variant of the dataset, ProtoNet base learner which is a nearest neighbor based works well because there is overlap in the speakers, but for the harder No-SPO variant, discriminating base learners SVM work better.

\begin{figure}[!tbp]
\centering
\begin{tabular}{cccc}
   \includegraphics[height=28mm,width=30mm]{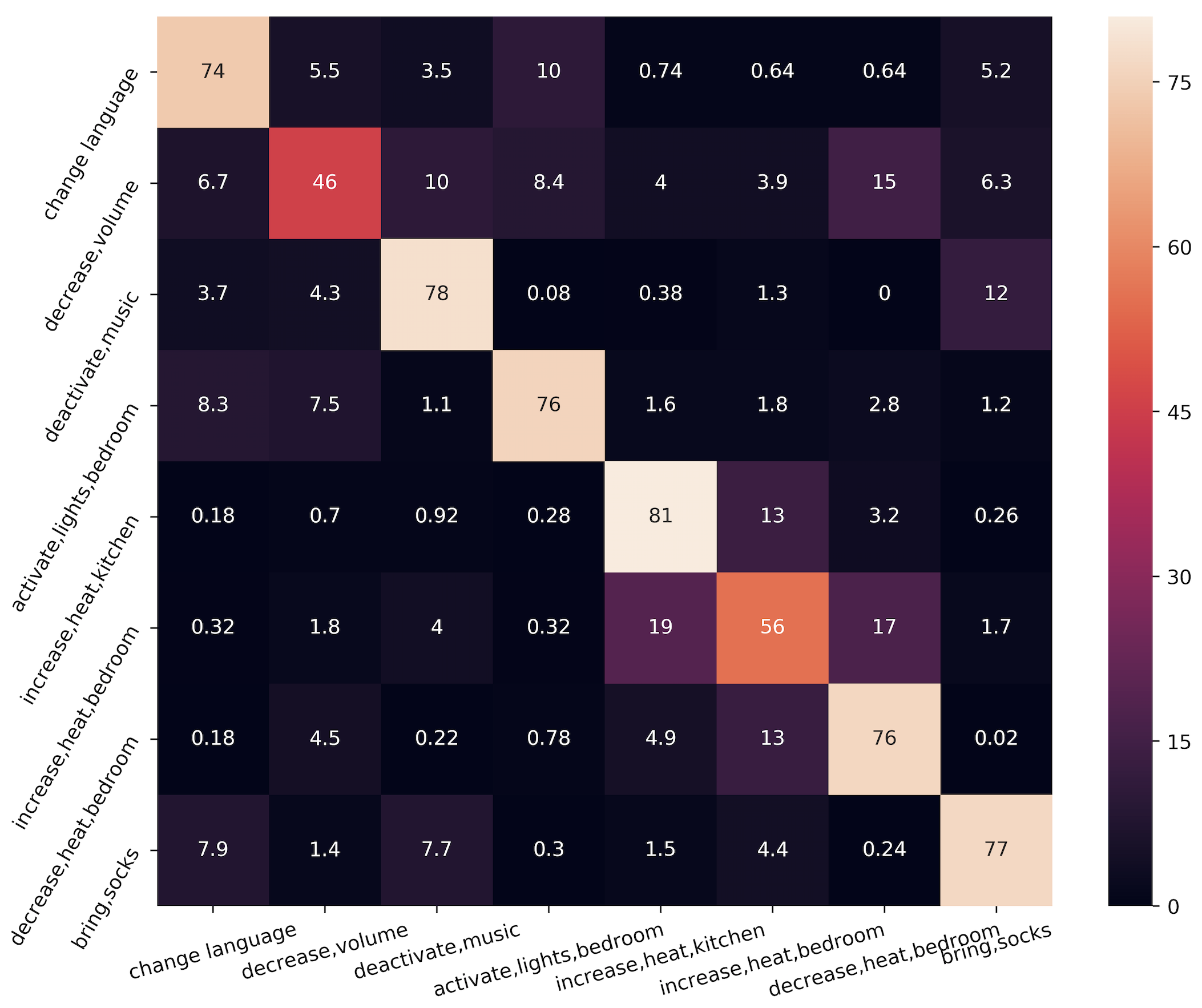}&
   \includegraphics[height=28mm,width=30mm]{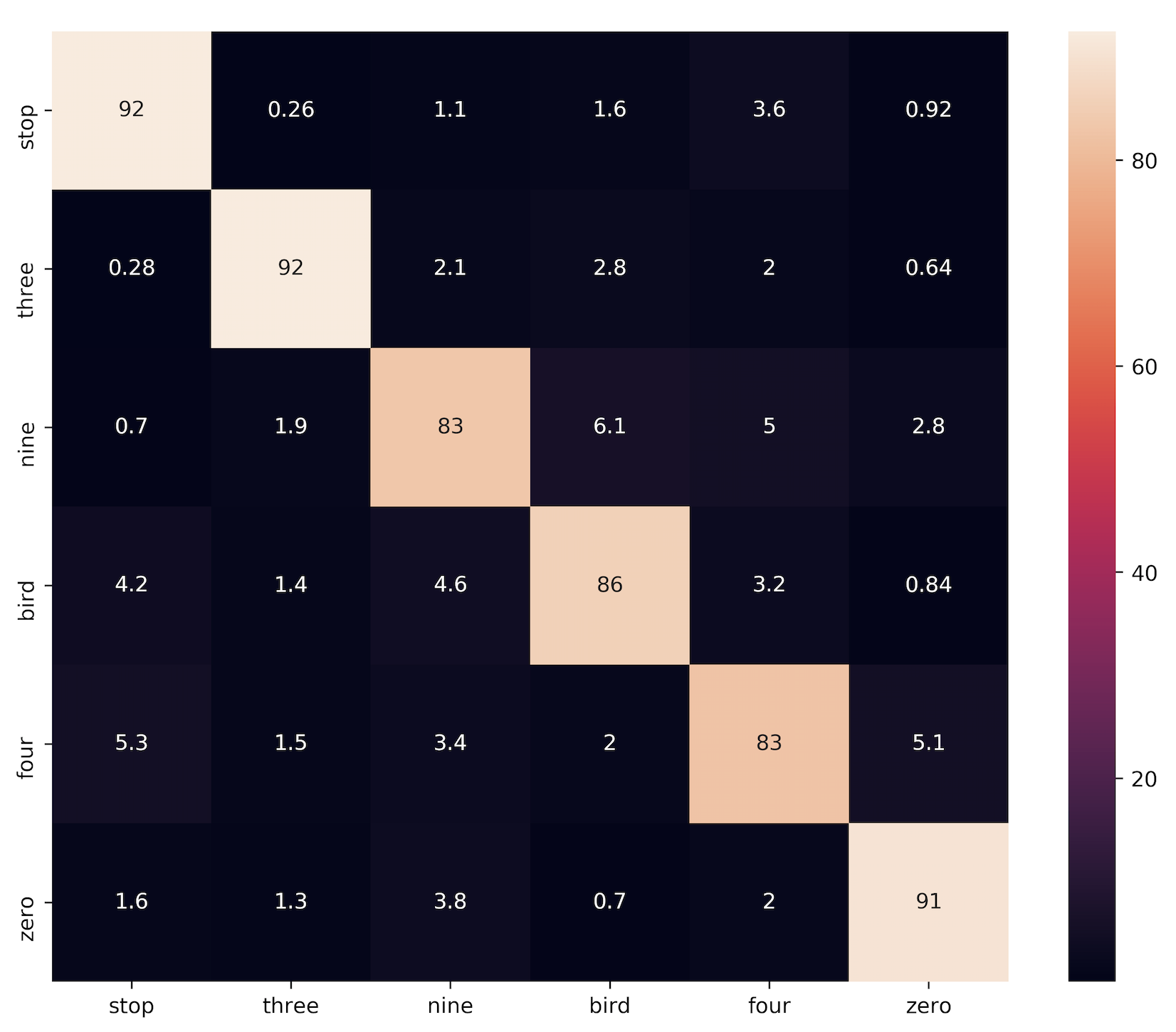}
   \\
(a) Fluent Speech Commands &
(b) Google Commands&
   \end{tabular}
\caption{The confusion matrix is mean computed for 5-shot classification task over 1000 test episodes.}
\label{fig:heat_map} 
\end{figure}

Generally, max-margin classifiers such as SVM tend to outperform nearest neighbor distance based approaches, we find that the performance of ProtoNet is often superior to SVM. Similarly, a closed-form solution of Ridge may be more sensitive to noise than ProtoNet leading to lower performance. Specifically, during the fine-tune process the linear SVM hyper parameters are fixed as (C=0.1, maxIter=15). The robustness of nearest neighbor, particularly in 1-shot makes the case for simpler models. Despite our exhaustive experimentation, we acknowledge that further hyper-parameter tuning may improve MetaOptNet.

\noindent Finally, Figure \ref{fig:heat_map} shows the confusion matrix for both the datasets on \emph{all} their test classes, i.e, 5-shot $n$-way with $n=6,8$ respectively. Notice that the model predictably fails on those intent classes that have significant overlap in their phrasings. In Fluent Speech Commands dataset, the class \textit{increase,heat,bedroom} is confused with the class \textit{increase,heat,kitchen} $19\%$ of the times and $17\%$ of the times with the class \textit{decrease,heat,bedroom}, all of which share lexicons with the each other. On the Google Commands dataset there is little overlap in lexicon thus the model learns robust discriminating features using few-shots as evident by the accuracy scores.

